%% file: main.tex
\documentclass[runningheads]{llncs}

\usepackage{eccv}

\usepackage{eccvabbrv}

\usepackage{graphicx}
\usepackage{booktabs}
\usepackage{multirow}
\usepackage{caption}
\usepackage{enumitem}
\usepackage{amsmath}
\usepackage{tabularx} 
\usepackage{arydshln}
\usepackage{makecell}
\usepackage{marvosym}

\usepackage[accsupp]{axessibility}  

\newcommand \footnoteONLYtext[1]{
	\let \mybackup \thefootnote
	\let \thefootnote \relax
	\footnotetext{#1}
	\let \thefootnote \mybackup
	\let \mybackup \imareallyundefinedcommand
}

\usepackage[pagebackref,breaklinks,colorlinks,citecolor=eccvblue]{hyperref}

\usepackage{orcidlink}

\begin{document}

\title{PreSight: Enhancing Autonomous Vehicle Perception with City-Scale NeRF Priors} 

\titlerunning{PreSight}

\author{Tianyuan Yuan\inst{1}\and
Yucheng Mao\inst{1}\text{$^{,}$}\inst{2}\and
Jiawei Yang\inst{3}\and
Yicheng Liu\inst{1}\and \\
Yue Wang\inst{3}\and
Hang Zhao\inst{1,4}\textsuperscript{\Letter}
}

\footnoteONLYtext{\Letter: Corresponding Author.}

\authorrunning{T.~Yuan et al.}

\institute{
    \textsuperscript{1}IIIS, Tsinghua University
    \textsuperscript{2}University of Science and Technology Beijing \\
    \textsuperscript{3}University of Southern California \hspace{1em} 
    \textsuperscript{4}Shanghai Qi Zhi Institute \\
    \email{\{yuanty22@mails, hangzhao@mail\}.tsinghua.edu.cn}}

\maketitle

\begin{abstract}
  Autonomous vehicles rely extensively on perception systems to navigate and interpret their surroundings. Despite significant advancements in these systems recently, challenges persist under conditions like occlusion, extreme lighting, or in unfamiliar urban areas. Unlike these systems, humans do not solely depend on immediate observations to perceive the environment. In navigating new cities, humans gradually develop a preliminary mental map to supplement real-time perception during subsequent visits.
  Inspired by this human approach, we introduce a novel framework, \textbf{PreSight}, that leverages past traversals to construct static prior memories, enhancing online perception in later navigations. Our method involves optimizing a city-scale neural radiance field with data from previous journeys to generate neural priors. These priors, rich in semantic and geometric details, are derived without manual annotations and can seamlessly augment various state-of-the-art perception models, improving their efficacy with minimal additional computational cost.
  Experimental results on the nuScenes dataset demonstrate the framework's high compatibility with diverse online perception models. Specifically, it shows remarkable improvements in HD-map construction and occupancy prediction tasks, highlighting its potential as a new perception framework for autonomous driving systems. Our code will be released at \url{https://github.com/yuantianyuan01/PreSight}.
  
\end{abstract}

\section{Introduction}
\label{sec:intro}
Vision-based perception modules serve as the cornerstone of autonomous vehicle navigation, ensuring the safety and reliability of these systems. These modules support a variety of perception tasks, such as 3D object detection~\cite{wang2021fcos3d, wang2022detr3d, li2022bevformer, huang2021bevdet, huang2022bevdet4d, lin2023sparse4d}, online map construction~\cite{li2021hdmapnet, liu2023vectormapnet, MapTR, Yuan_2024_streammapnet, xiong2023neuralmapprior}, and occupancy prediction~\cite{huang2023tri, tian2023occ3d, li2023voxelformer, li2023voxformer}, each focusing on different aspects of the vehicle's surroundings. Perception tasks can be broadly classified into dynamic object perception, which deals with moving entities like vehicles and pedestrians, and static environment perception, focusing on invariant features such as lane markers, vegetation, and obstacles. While dynamic objects require real-time observation, static environments can be pre-mapped and annotated, enhancing navigational safety. 
Conventionally, static environments are captured in high-definition (HD) maps using Simultaneous Localization and Mapping (SLAM) techniques~\cite{legoloam2018, loamzhang2014, segal2009generalized, dellaert2012factor}, supplemented with manual annotations. Despite their precision, expanding these methods to cover new geographic regions or incorporate new attributes presents considerable challenges, primarily due to the extensive reliance on manual labor for annotations and updates. In response, online perception models powered by deep learning have been developed~\cite{li2021hdmapnet, liu2023vectormapnet, MapTR, li2023fbocc}, utilizing only onboard sensor data to infer the surrounding environment. However, these models often struggle with occlusions, adverse weather, or extreme lighting conditions, and when deployed in unfamiliar urban settings not included in their training datasets, their reliability is compromised~\cite{Yuan_2024_streammapnet, lilja2023localization}.

\begin{figure}[tb]
  \centering
  \includegraphics[width=1.0\linewidth]{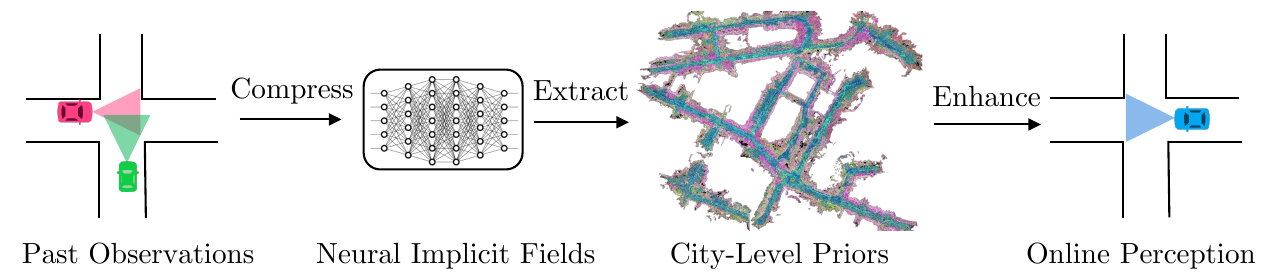}
  \caption{Pipeline of PreSight. Leveraging historical traversal data, PreSight enhances online perception through NeRF, compressing extensive observations into implicit fields. A ray-marching algorithm extracts structured city-level priors, facilitating improved perception in subsequent visits.
  }
  \label{fig:main_teaser}
\end{figure}
Consider the innate human capability to navigate through cities. Individuals inherently develop preliminary mental maps of urban landscapes, with varying degrees of accuracy. These priors serve as valuable supplements to real-time sensory perception during subsequent explorations. Inspired by this human capability, we propose a novel framework, \textbf{PreSight}, illustrated in \Cref{fig:main_teaser}, that utilizes historical traversal data—gathered through crowdsourcing or personal journeys—to automatically generate static priors without manual annotation efforts. These unsupervised, semantically and geometrically rich priors significantly enhance the efficacy of online perception models in subsequent navigations.
Our framework includes city-scale neural radiance fields (NeRF) ~\cite{mildenhall2020nerf} priors and a straightforward integration module for various online perception models. Initially, we employ city-scale NeRFs to reconstruct urban environments using camera images and poses captured from past traversals, without requiring LiDAR input. Given the extensive spatial expanse of these cities, we divide them into tiles, each encompassing approximately $1~km^2$. For each tile, several Instant-NGP~\cite{mueller2022instant} sub-fields are deployed and optimized in parallel for detailed coverage. Semantic information is distilled into the NeRFs by leveraging features from pre-trained 2D visual foundation models with semantic supervision. Structured priors are then extracted using a ray marching algorithm inspired by EmerNeRF~\cite{yang2023emernerf}, selecting occupied voxels and gathering their semantic features from the constructed NeRFs. These featurized prior voxels serve as robust supplements to online perception.
To integrate with prevalent online models~\cite{li2023fbocc, huang2022bevdet4d, MapTR, Yuan_2024_streammapnet}, we encode these priors into BEV or 3D features, using straightforward CNNs to merge features from both priors and real-time observations. The details of each part of our framework are provided in \Cref{sec:methodologies}.

We validate our framework on the widely used nuScenes~\cite{caesar2020nuscenes} dataset, focusing on two important online perception tasks: local HD-map construction and occupancy prediction. To illustrate our method's efficacy in enhancing the adaptability of online perception models to novel environments—and to avoid data leakage—we re-split the training-testing set to eliminate any geometric overlaps. The experimental results demonstrate that our framework significantly boosts the performance of several perception models across both tasks. For an in-depth discussion on the dataset and the experimental setup, please see \Cref{sec:experiments}.

To summarize, our contributions are as follows:
\begin{itemize}
    \item We introduce a novel framework, \textbf{PreSight}, that leverages historical data to automatically construct powerful priors using NeRF.
    \item We demonstrate that the generated priors can be seamlessly integrated into various perception models and tasks, significantly enhancing their performance with minimal modification.
\end{itemize}

\section{Related Work}
\label{sec:related_work}
\subsection{Conventional 3D Reconstruction Techniques}
Various 3D reconstruction techniques have been developed to create 3D models from 2D images.
Structure from Motion (SfM) uses the relative motion between different views in a set of 2D images to reconstruct 3D structures~\cite{ransac, lowe2004distinctive}. It is commonly applied for estimating camera positions and generating 3D point clouds of scenes. However, SfM typically produces sparse and coarse geometric reconstructions. It also faces challenges in complex scenarios with occlusions~\cite{2012arie_global, phototourism, Crandall2011Discrete}.
Visual SLAM is regarded as a specialized variant of the SfM problem, tailored for robotics and autonomous driving applications. By incorporating pose information from GPS or IMU, vSLAM achieves more precise camera pose estimation and better reconstruction quality. Nonetheless, it encounters similar limitations in dynamic, complex, and large-scale environments~\cite{Aulinas2008slam, Fuentes2015vslam, Goldstein2016ShapeFit, Özyeşil_Voroninski_Basri_Singer_2017}. Additionally, infusing semantic information into the reconstructed models is not inherently straightforward with conventional 3D reconstruction techniques.

\subsection{Large-Scale Neural Radiance Fields}
The application of Neural Radiance Field (NeRF) in urban scenarios has attracted significant interest. Block NeRF~\cite{tancik2022blocknerf} attempts to reconstruct city-scale scenes using traditional MLP-based NeRF from densely captured camera images, but suffers from prolonged training durations. SUDS~\cite{turki2023suds} employs hash grids~\cite{mueller2022instant} for faster training and divides large areas into multiple sub-fields for detailed coverage. StreetSurf~\cite{guo2023streetsurf}, focusing on static scenarios in autonomous driving, delivers high-quality novel view synthesis and geometry but overlooks dynamic objects. 
UniSim~\cite{yang2023unisim} and NeuRAD\cite{neurad} build neural simulators using NeRFs, and use SDF to model the surfaces.
EmerNeRF~\cite{yang2023emernerf} distinguishes between dynamic and static elements in a self-supervised manner and leverages vision foundation models for knowledge distillation. However, these approaches often depend on LiDAR depth data or dense observational views, limiting scalability in new regions via crowdsourcing. OccNeRF~\cite{chubin2023occnerf} and SelfOcc~\cite{huang2023self} explore using NeRF for autonomous driving scene occupancy estimation but neglect dynamic entities and underperform compared to online perception models, questioning their suitability as ground truth. MV-Map~\cite{xie2023mvmap} leverages NeRF for offboard HD-map generation.
Our methodology, in contrast, focuses on generating priors to enhance various online perception models.

\subsection{Priors by Multi-Traversals}
\label{sec:neural_map_priors}
Recent work has explored on the nuScenes dataset~\cite{caesar2020nuscenes} to enhance online perception in autonomous driving using priors from previous traversals. HindSight~\cite{you2022hindsight} utilizes LiDAR point clouds from past journeys to aid online detection tasks and proposes to re-split the nuScenes dataset to provide multi-traversal samples without geometrical overlaps, mitigating data leakage risks. Yet, these historical point clouds lack semantic depth, narrowing their utility. Neural Map Prior~\cite{xiong2023neuralmapprior} employs neural representations to build and update a global map prior, boosting online map perception. Nonetheless, the substantial memory requirement for its neural representation limits scalability. Both approaches, while innovative, are confined to particular perception tasks and underscore the need for versatile priors that can support a broader range of tasks.

\subsection{Visual Online Perception Model}
Visual online perception models leverage camera image input to predict the surrounding environment in real-time, including a range of tasks from 3D object detection~\cite{li2022bevformer, huang2021bevdet, huang2022bevdet4d, Yang2022BEVFormerVA, liu2022bevfusion, li2023fbbev} to online HD map construction~\cite{Yuan_2024_streammapnet, liu2023vectormapnet, li2021hdmapnet, MapTR, ding2023pivotnet} and occupancy prediction~\cite{huang2022bevdet4d, li2023fbocc, hong2024univision}. 
Particularly, we focus on online HD map construction and occupancy prediction. Online HD map construction is crucial for autonomous navigation, utilizing onboard sensors to create local HD maps. These maps can be vectorized~\cite{liu2023vectormapnet, MapTR, Yuan_2024_streammapnet} or rasterized~\cite{roddick2020predicting, xiong2023neuralmapprior}, providing detailed semantic guidance to ensure vehicular safety. This approach offers a cost-effective, scalable alternative to traditional, manually annotated HD maps by minimizing the need for human labor.
The occupancy prediction task, pioneered by Occ3D~\cite{tian2023occ3d}, involves predicting the occupancy and semantic label of each voxel in a scene from camera images. This task delivers comprehensive information about the vehicle’s surrounding environment but presents significant challenges in achieving accurate geometric and semantic predictions.

\section{Methodologies}
\label{sec:methodologies}
\begin{figure}[tb]
  \centering
  \includegraphics[width=1.0\linewidth]{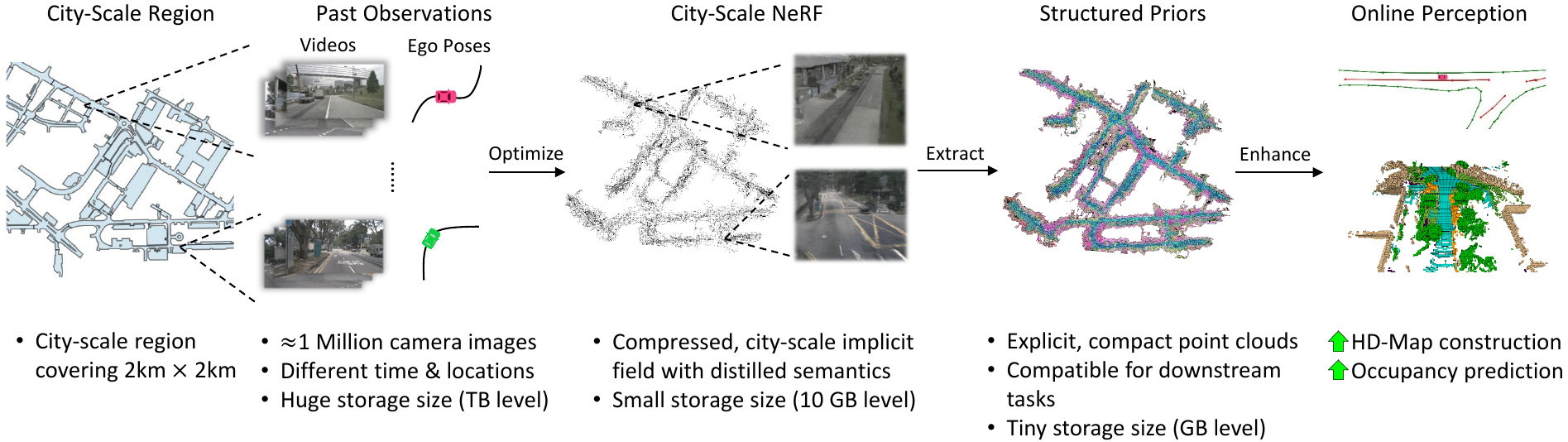}
  \caption{A detailed pipeline of PreSight. It starts by optimizing city-scale NeRFs from past traversals' observation. Then structured priors are extracted from NeRFs to enhance online perception models. 
  }
  \label{fig:teaser2}
\end{figure}
Our framework leverages self-supervised Neural Radiance Fields (NeRFs)~\cite{mildenhall2020nerf} to generate static city priors from historical data and integrates these priors into online perception models, as demonstrated in~\Cref{fig:teaser2}. This section is structured as follows: We begin by defining the problem in Section \ref{sec:problem_formulation}, proceed to describe the construction of city-scale NeRFs fused with semantic features in Section \ref{sec:city_scale_nerf}, introduce the extraction of structured priors in Section \ref{sec:prior_extraction}, and conclude with our approach to integrating these priors with state-of-the-art online perception models in Section \ref{sec:integrating}.

\subsection{Problem Formulation}
\label{sec:problem_formulation}
Consider a set of training samples $X_{train}$ and their corresponding annotations $Y_{train}$. Typically, an online perception model $M$ is trained on the dataset \\ $\left(X_{train}, Y_{train}\right)$. At deployment, this model is expected to effectively generalize to unseen samples $X_{test}$, including those from entirely new locations—a scenario where many advanced perception models encounter difficulties. Suppose we also have access to a set of unlabeled samples $X_{prior}$, which geographically overlap with $X_{test}$ and can be readily acquired through crowdsourcing. Our objective is to effectively utilize the unlabeled $X_{prior}$ to autonomously generate priors $P_{prior}$, thereby enhancing the performance of model $M$ on $X_{test}$.

More specifically, during the training phase, we generate priors $P_{train}$ based on $X_{train}$. An enhanced perception model $M'$, equipped with an integration module, is then trained on the combined dataset $\left(X_{train} + P_{train}, Y_{train}\right)$. Before evaluation, we generate priors $P_{prior}$ using the unlabeled $X_{prior}$. The model $M'$ is subsequently evaluated on the dataset $\left(X_{test} + P_{prior}, Y_{test}\right)$. It is crucial to note that $X_{train}$ is geographically distinct from $X_{test} \cup X_{prior}$ to prevent data leakage. For more details about the experimental setup, please refer to \Cref{sec:setup}.

\subsection{Building City-Scale NeRF}
\label{sec:city_scale_nerf}
\textbf{Inputs.} Our method begins by taking sequences of RGB images from videos and their associated camera poses. To enhance these inputs, we utilize pre-trained vision foundation models for generating additional descriptors on pixels. Specifically, we employ DINO~\cite{caron2021emerging} for camera-view semantic feature generation and SegFormer~\cite{xie2021segformer} for producing camera-view semantic segmentation masks.

\noindent\textbf{Benefits of Using NeRF in Our Framework.}
Neural Radiance Fields (NeRF) \cite{mueller2022instant} excel at generating continuous, consistent, and high-fidelity 3D representations from sparse and irregularly spaced images, setting them apart from traditional 3D reconstruction techniques. Unlike methods that depend on specialized inputs such as LiDAR data or stereo camera imagery, NeRF operates effectively with standard 2D images. 
Moreover, NeRF's volumetric rendering approach allows for the straightforward exclusion of dynamic objects from scenes and the integration of pixel-wise semantic features~\cite{yang2023emernerf}. These features can be extracted from powerful vision foundation models and seamlessly incorporated into the 3D field. In our framework, we use NeRF not merely as a tool for synthesizing novel views but as an efficient, self-supervised learner of environmental priors. Our focus is on the quality of geometric and semantic reconstructions rather than on the fidelity of novel view generation, leveraging NeRF's strengths to enhance the robustness and accuracy of online perception models in static environment tasks.

\begin{figure}[tb]
  \centering
  \begin{subfigure}{0.73\linewidth}
    \includegraphics[width=1.0\linewidth]{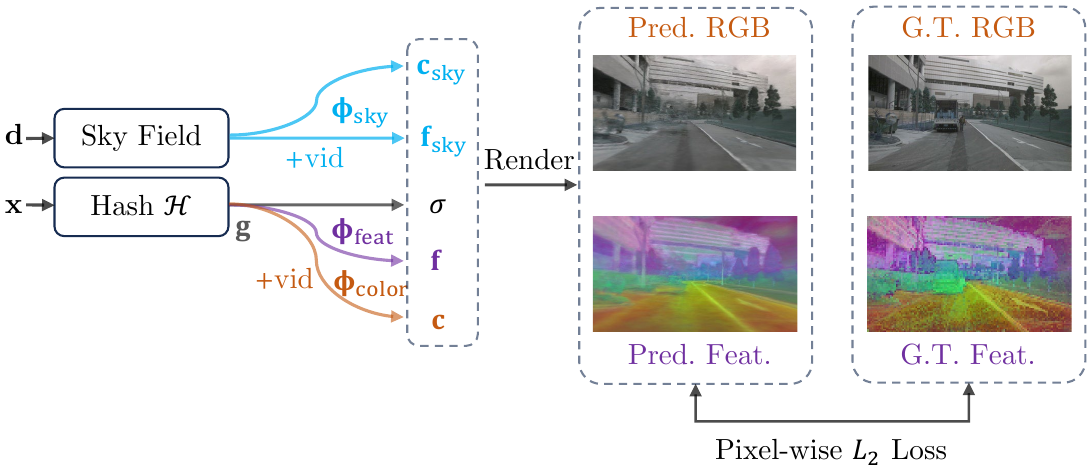}
    \caption{PreSight's NeRF constructs distinct fields for nearby scenes and the sky, utilizing video embeddings to accommodate variations in lighting conditions across scenes. Semantic features are merged similar RGB color blending, enabling the distillation of semantic knowledge from DINO~\cite{caron2021emerging}.}
    \label{fig:short-a}
  \end{subfigure}
  \hfill
  \begin{subfigure}{0.25\linewidth}
    \includegraphics[width=0.85\linewidth]{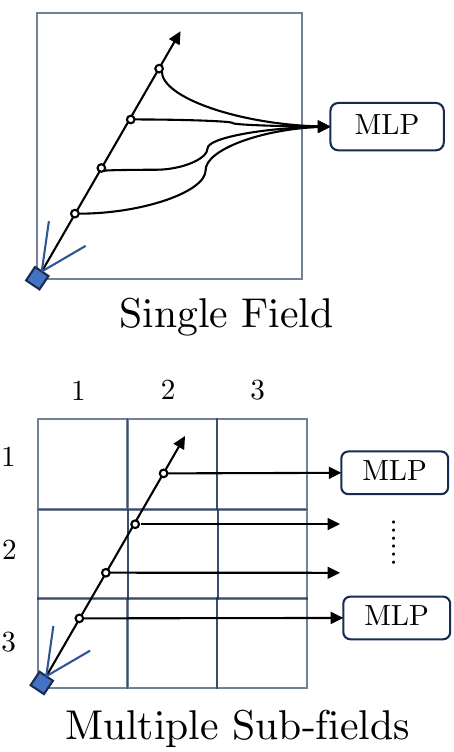}
    \caption{Multiple sub-fields are used to guarantee comprehensive coverage. Volumetric rendering directly integrates sample points across these sub-fields.}
    \label{fig:short-b}
  \end{subfigure}
  \caption{Overview of PreSight's scene representation.}
  \label{fig:short}
\end{figure}
\noindent\textbf{Scene Representation.}
Given the vast spatial extent of input videos throughout the city, we partition the city into tiles, each approximately $1~km^2$. Each tile is independently optimized using video data overlapping with its designated area. We apply a straightforward application of K-Means clustering on all camera poses to determine tile boundaries.

For each tile, we construct a large static scene representation consisting of multiple sub-fields ${\mathbf{F}^i}$ to ensure comprehensive coverage. The query outputs are aggregated from these sub-fields along the ray during the rendering process. Similar to the initial clustering, we employ K-Means clustering on camera poses to establish centroids for these sub-fields. Each sub-field mimics the structure found in Instant-NGP~\cite{mueller2022instant}, utilizing multi-resolution hash grids $\mathcal{H}$ coupled with a small MLP $g$ to encode the implicit spatial features. Separate heads $\phi_\mathrm{color}, \phi_\mathrm{feat}, \phi_\mathrm{sky}$ are utilized for predictions of color, semantic information, and sky appearance, respectively. The queried inputs include $(\mathbf{x}, \mathbf{d}, vid)$, where $\mathbf{x}$ represents the 3D location, $\mathbf{d}$ the view direction, and $vid$ the video identifier:
\begin{equation}
  \mathbf{g}, \sigma = g\left( \mathcal{H}(\mathbf{x}) \right)
\end{equation}

\noindent $\mathbf{g}, \sigma$ represent the per-point feature and density, respectively. For color predictions:
\begin{equation}
    \mathbf{c} = \phi_\mathrm{color}\left(\mathbf{g}, \gamma(\mathbf{d}), \mathbf{V}(vid)\right)
\end{equation}
\noindent $\gamma(\cdot)$ denotes spherical harmonic encoding for direction, and $\mathbf{V}(\cdot)$ represents per-video embeddings to account for video-specific lighting conditions. For semantic feature predictions:
\begin{equation}
    \mathbf{f} = \phi_\mathrm{feat}\left(\mathbf{g}\right)
    \label{eq:feat}
\end{equation}
\noindent This process models direction-invariant semantic features. Overall, the output for each query is given by:
\begin{equation}
    \sigma, \mathbf{c}, \mathbf{f} = \mathbf{F}(\mathbf{x}, \mathbf{d}, vid)
\end{equation}
For sky features, a position-independent MLP $\phi_{sky}$ predicts both color and semantic features:
\begin{equation}
\mathbf{c}_\mathrm{sky}(\mathbf{d}), \mathbf{f}_\mathrm{sky}(\mathbf{d}) = \phi_\mathrm{sky}\left(\gamma(\mathbf{d}), \mathbf{V}(vid)\right)
\end{equation}

\noindent\textbf{Rendering.}
To render a camera ray $\mathbf{r}=(\mathbf{o}, \mathbf{d})$ within a video $vid$ (where $o$ denotes the origin), we sample $N$ points $\left\{\mathbf{x}^i\right\}$ along the direction $\mathbf{d}$. Each sample point is assigned to the nearest sub-field. The assignment process is defined as:
\begin{equation}
    \delta (\mathbf{x}^i) = \mathop{\arg\min}\limits_{j} \left\Vert \mathbf{x}^i - \mathbf{centroid}^j \right\Vert_2
\end{equation}
where $\mathbf{centroid}^j$ denotes the centroid of $j$-th sub-field. After assignment, outputs for each sample point are queried exclusively from the assigned sub-field:
\begin{equation}
    \sigma^i, \mathbf{c}^i, \mathbf{f}^i = \mathbf{F}^{\delta(\mathbf{x}^i)} (\mathbf{x}^i, \mathbf{d}, vid)
\end{equation}

\noindent The final rendering is an integration along the ray, blending color and semantic features:
\begin{equation}
    \hat{C}(\mathbf{r}) = \sum_{i=1}^N T^i \alpha^i \mathbf{c}^i + \left(1 - \sum_{i=1}^N T^i \alpha^i\right)\ \mathbf{c}_\mathrm{sky}(\mathbf{d})
\end{equation}
\vspace{-1ex}
\begin{equation}
    \hat{F}(\mathbf{r}) = \sum_{i=1}^N T^i \alpha^i \mathbf{f}^i + \left(1 - \sum_{i=1}^N T^i \alpha^i\right)\ \mathbf{f}_\mathrm{sky}(\mathbf{d})
\end{equation}
$\alpha^i$ and $T^i$ denote the accumulated transmittance and piece-wise opacity, respectively, calculated as follows:
\begin{equation}
    \alpha^i=1-\exp \left(-\sigma^i(\mathbf{x}^{i+1}-\mathbf{x}^i)\right)
    \label{eq:weights1}
\end{equation}
\vspace{-1ex}
\begin{equation}
    T^i=\prod_{j=1}^{i-1}(1-\alpha^j)
    \label{eq:weights2}
\end{equation}

\noindent\textbf{Optimization.} 
We employ an L2 loss for color $\mathcal{L}_\mathrm{rgb}$ and semantic feature reconstruction $\mathcal{L}_\mathrm{feat}$, along with a binary cross-entropy loss for sky features $\mathcal{L}_\mathrm{sky}$ following approach introduced in~\cite{yang2023emernerf, guo2023streetsurf}. Inspired by Mip-NeRF 360~\cite{barron2022mipnerf360}, we use two proposal networks, integrating interlevel and distortion losses. The comprehensive loss function is represented as:
\begin{equation}
    \mathcal{L} = \mathcal{L}_\mathrm{rgb} + \lambda_\mathrm{feat}\mathcal{L}_\mathrm{feat} + \lambda_\mathrm{sky}\mathcal{L}_\mathrm{sky} + \lambda_\mathrm{inter}\mathcal{L}_\mathrm{inter} + \lambda_\mathrm{dist}\mathcal{L}_\mathrm{dist}
\end{equation}
To focus on static environment features, we mask out pixels associated with potentially moving objects, such as vehicles and pedestrians, using segmentation masks derived from SegFormer~\cite{xie2021segformer}.

\subsection{Prior Extraction}
\label{sec:prior_extraction}
After the optimization phase, the NeRFs encapsulate rich information within their hash grids and MLPs. To make this dense, unstructured information inherent in NeRFs usable for online perception models, it requires conversion into a structured form. We employ a ray marching algorithm to identify the voxels that are occupied and to capture their associated features. The process begins by casting rays originating from the training cameras. Similar to the training procedure, we sample $N$ sample points $\left\{\mathbf{x}^i\right\}$ along each ray. The surface point is determined by locating the first point where the cumulative transmittance and opacity exceed a threshold:
\begin{equation}
    j = \min_i \left( \sum_{i=1}^{N} T^i\alpha^i > 0.5 \right)
\end{equation}
Here, $T^i$ and $\alpha^i$ are defined in \Cref{eq:weights1} and \Cref{eq:weights2}. We collect the identified surface point $\mathbf{x}^j$ and its corresponding semantic feature $\phi_\mathrm{feat}\left(\mathbf{g}\right)$, as previously defined in \Cref{eq:feat}. 
Following the aggregation of surface points from all training views, we apply voxel-based downsampling to minimize the points, averaging the features within each voxel. 
These feature-enriched voxels then act as robust priors for enhancing online perception models. It is worth noting that the extraction is only performed once after building the city-scale NeRF, and the extracted priors are stored. Neither the slow rendering process or the huge size of NeRFs impact the online perception models. During deployment, only the extracted priors are stored and used by the online perception models.

\begin{figure}[tb]
  \centering
  \includegraphics[width=0.85\linewidth]{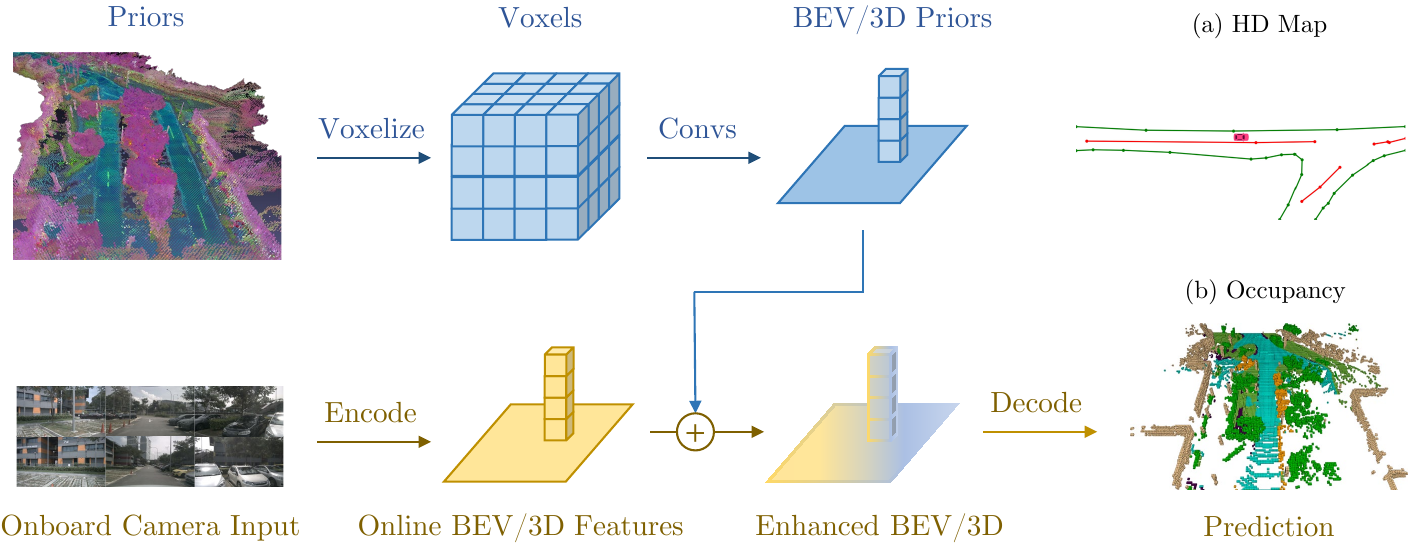}
  \caption{The integration module. Point-based priors are voxelized and then encoded into BEV or 3D features using convolutional layers. These features are fused with online features from perception models, enhancing the overall features for decoding.
  }
  \label{fig:integration}
\end{figure}
\subsection{Integrating Priors to Online Models}
\label{sec:integrating}
Online perception models commonly transform image inputs into a canonical representation, such as BEV or 3D space features. To ensure our framework's broad compatibility, we adapt our pre-built priors to BEV or 3D space, simplifying their integration with various state-of-the-art online perception models~\cite{li2023fbocc, huang2022bevdet4d, Yuan_2024_streammapnet, MapTR}. As illustrated in \Cref{fig:integration}, our integration strategy employs several simple CNN layers to transform the voxel features into BEV or 3D space, which are then concatenated with the existing online features from the perception models. 
This module introduces negligible computational overhead and requires no alterations to the original model architectures, efficiently enhancing the models with rich environmental priors for improved real-time perception capabilities.

\section{Experiments}
\label{sec:experiments}
In this section, we evaluate our \textbf{PreSight} framework on the nuScenes~\cite{caesar2020nuscenes} dataset, specifically targeting perception tasks involving static targets: online HD map construction and occupancy prediction. These tasks are chosen because our framework is designed to leverage priors to enhance static environment perception, a crucial part for these applications. The dataset are introduced in \Cref{sec:setup}, followed by NeRF implementation details in \Cref{sec:nerf_implementation}. \Cref{sec:results} showcases our framework's enhancements to baseline models. Our priors are compared to alternatives in \Cref{sec:comparison}, and \Cref{sec:ablation} presents some ablation studies of our methods.

\subsection{Setup}
\label{sec:setup}
We evaluate our framework using the nuScenes~\cite{caesar2020nuscenes} dataset, a leading benchmark in autonomous driving, collected across four regions: \textit{Boston Seaport}, \textit{Singapore Onenorth}, \textit{Singapore Queenstown}, and \textit{Singapore Holland Village}, featuring 1000 scenes with six synchronized cameras and vehicle poses. The official dataset split allocates 700 scenes for training and 150 for validation. However, recent studies have identified data leakage issues within this official split, highlighting the need for a revised partitioning to accurately assess static perception models' generalization capabilities. 
Following our problem setup in \Cref{sec:problem_formulation}, we assign scenes from \textit{Boston Seaport} and \textit{Singapore Queenstown} to our training set $\left(X_{train}, Y_{train}\right)$. For testing, one-third of the scenes from \textit{Singapore Onenorth} and \textit{Singapore Holland Village} are used as $X_{prior}$ for generating priors, with the remaining two-thirds forming the testing set $\left(X_{test}, Y_{test}\right)$.
Therefore, following the problem setup detailed in \Cref{sec:problem_formulation}, we set all samples in \textit{Boston Seaport} and \textit{Singapore Queenstown} as training set $\left(X_{train}, Y_{train}\right)$. 
This distribution strategy mirrors the practical scenario of scaling and generalizing an autonomous driving system: training is conducted using data from specific cities, followed by deployment in previously unencountered cities, supported by a limited set of observations $X_{prior}$ sourced through crowdsourcing.
Overall, we established a split of 614 training scenes, 82 prior scenes, and 154 testing scenes, with an $80\%$ geographical overlap between the prior and testing sets. Further details on scene allocation can be found in the supplemental material.

\subsection{NeRF Implementation}
\label{sec:nerf_implementation}
This section introduces the implementation details for constructing city-scale Neural Radiance Fields (NeRFs). We utilize 12Hz image sequences from six cameras provided by the nuScenes dataset, yielding approximately 1,400 images for each 20-second scene. following the approach of EmerNeRF\cite{yang2023emernerf}, we compress the dimension of DINO features from 764 to 64. Dynamic objects including \textit{person, rider, car, truck, bus, train, motorcycle,} and \textit{bicycle} are masked using segmentation results from SegFormer\cite{xie2021segformer}.
As detailed in \Cref{sec:city_scale_nerf}, urban areas are divided into tiles of roughly $1~km^2$. Specifically, we assign 8, 4, 4, and 2 tiles to \textit{Boston Seaport}, \textit{Onenorth}, \textit{Queenstown}, and \textit{Holland Village}, respectively, based on their sizes. Each tile contains 8 or 16 sub-fields, depending on the tile's area. NeRF training for each tile involves 100,000 iterations with a batch size of $2^{16}$, requiring about 20 hours on a single A100 GPU.

\subsection{Experimental Results on Online models}
\label{sec:results}
\input{tables/vector_map_1_runtime}
\noindent\textbf{Online HD Map construction.} 
Online HD maps construction takes as input camera images to generate local HD maps, which can be either vectorized~\cite{MapTR, Yuan_2024_streammapnet} or rasterized~\cite{xiong2023neuralmapprior}. Our research demonstrates that the integration of proposed priors enhances the performance of state-of-the-art models. Specifically, for vectorized mapping, we employed two models with distinct decoders, namely MapTR~\cite{MapTR} and StreamMapNet~\cite{Yuan_2024_streammapnet}. For rasterized mapping, we adopted a BEVFormer-based~\cite{li2022bevformer} model, following Neural Map Prior~\cite{xiong2023neuralmapprior}. These models utilize BEV features as an intermediate representation, which are fused with our pre-built priors using our integration module. We trained and evaluated each model based on their original codebase, with our proposed training-testing split. The perception range was set to $100\times50$ meters, aligning with StreamMapNet's specifications. Given that map construction is inherently a static perception task, our framework significantly elevates the quality of the resulting HD maps across all models \Cref{tab:vector_map_1}.
The performance metrics for the baseline models are lower than those originally reported in their respective papers due to our use of a more challenging dataset split without geographical overlaps, a phenomenon similarly observed by Lilja \etal~\cite{lilja2023localization} and Yuan \etal~\cite{Yuan_2024_streammapnet} More implementation details are provided in the supplemental materials.

\input{tables/occ_1_runtime}
\noindent\textbf{3D Occupancy Prediction.}
3D occupancy prediction task involves estimating both the occupancy state and semantic label of each voxel in a scene, using camera images as input. We benchmark our approach using the widely recognized Occ3D dataset~\cite{tian2023occ3d} on nuScenes. For baselines, we selected FB-OCC~\cite{li2023fbocc} and BEVDet~\cite{huang2022bevdet4d}, two leading state-of-the-art models with distinct architectures, integrating our pre-built priors into their BEV features. Each model was trained and evaluated on their original codebase, with our proposed training-testing split. The results, presented in \Cref{tab:occ_1}, demonstrate our framework's ability to enhance the baseline models' performance. Notably, we specifically highlight static categories such as \textit{drive surface}, \textit{other flat}, \textit{sidewalk}, \textit{terrain}, \textit{manmade}, and \textit{vegetation}. Our framework significantly improves outputs for these targets, without impacting dynamic categories, aligning with our goal of enhancing static perception accuracy.

\subsection{Comparison with Other Priors}
\label{sec:comparison}
This section compares our generated priors with alternative approaches to evaluate their effectiveness.

\noindent\textbf{Historical Point Clouds.}
A straightforward method for utilizing historical data involves collecting LiDAR-generated point clouds from previous traversals~\cite{you2022hindsight}. Despite our framework's independence from LiDAR, we assessed whether our priors could outperform these point clouds, known for their geometric precision. We replaced our priors with historical LiDAR point clouds to evaluate performance in HD map construction and occupancy prediction tasks.

\noindent\textbf{Neural Map Prior.}
Neural Map Prior~\cite{xiong2023neuralmapprior} enhances online map perception models with automatically generated priors, aligning with our objectives. We adapted this method to our experimental setup by training it with our dataset $\left(X_{train}, Y_{train}\right)$ and generating neural priors from $X_{prior}$ for evaluation on \\$\left(X_{test}, Y_{test}\right)$, making best use of its framework's potential. This comparison conducted on rasterized map construction task, following the official implementation. Notably, our priors are significantly more compact, requiring only 3.5 GB for \textit{Boston Seaport}, compared to the 38 GB needed for the rasterized BEV features in Neural Map Prior.

\input{tables/compare}
The results in \Cref{tab:compare} highlight our framework's superiority in both perception tasks. For HD map construction, our priors offer richer semantic details. In occupancy prediction, our approach matches historical LiDAR point clouds. While LiDAR slightly enhances static target perception, it fails to differentiate dynamic from static objects, causing potential inaccuracies. Our method maintains performance without compromising dynamic object perception.

\subsection{Ablation Studies}
\label{sec:ablation}
This section explores the impact of various components of our framework on the final outcomes. Specifically, we investigate the role of distilled DINO~\cite{caron2021emerging} features and the effect of varying the proportion of $X_{prior}$ utilized for constructing priors. 

\input{tables/nofeats}
\noindent\textbf{Distilled DINO Features.}
A crucial element of our framework is incorporating distilled semantic knowledge from the robust vision foundation model, DINO, into our priors. As demonstrated in \Cref{tab:no_feats}, removing these distilled semantic features leads to a notable decline in performance for both tasks, highlighting the importance of integrating distilled semantic knowledge. Particularly, HD map construction benefits greatly from semantic features for identifying map elements, whereas occupancy prediction depends more on geometrical information.

\input{tables/prior_ratio}
\noindent\textbf{Proportions of Used Prior Scenes.}
The effectiveness of our framework increases with the amount of historical traversal data utilized. To illustrate this, we evaluated the trained model using varying proportions of prior scenes. The outcomes are summarized in \Cref{tab:prior_ratio}. Performance improves as more scenes are utilized to create priors, showcasing our framework's scalability. Importantly, models trained with priors do not depend exclusively on them and still achieve performance comparable to baseline models with no priors at test time.

\subsection{Localization Errors}
In this section, we discuss the impact of localization errors on our method, focusing on two types:

\noindent\textbf{Pose errors in prior data before optimizing NeRFs.}
When building NeRFs offline, techniques like visual SLAM and visual odometry can reduce localization errors to under 10 cm, which our NeRFs can handle. For example, the NuScenes dataset lacks height information in ego-poses, resulting in about a 10 cm error in camera poses, yet our NeRFs manage this well.

\noindent\textbf{Localization noise during online perception integration.}
\input{tables/localization}
Autonomous vehicles using GNSS with IMU or RTK typically have pose errors of 0.3~m to 0.5~m. To test our models' robustness to localization noise, we added 2D Gaussian noise of $N(0, 0.5^2)$ to the ego-poses during evaluation. As shown in \Cref{tab:localization}, while localization noise slightly impacts performance, models with priors still outperform those without.

\subsection{Performance without Online Observation}
\label{sec:prior_only}
\input{tables/prior_only}
This section examines the performance of models using only priors, without online observations. We consider two settings: (a) Evaluating models trained with both online observations and priors, but replacing the online BEV features with zeros. (b) Training new models using only priors, ignoring the "encoding online observation" part during both training and testing, serving as "look-up" models or automatic map annotators.
Results in \Cref{tab:prior_only} show that models using only priors at test time perform poorly. However, models trained with priors only perform decently, indicating the potential for using priors to automatically annotate maps. Note that these results are evaluated where $X_{prior}$ and $X_{test}$ overlap, and online features are still necessary where priors do not cover.

\begin{figure}[tb]
  \centering
  \includegraphics[width=0.95\linewidth]{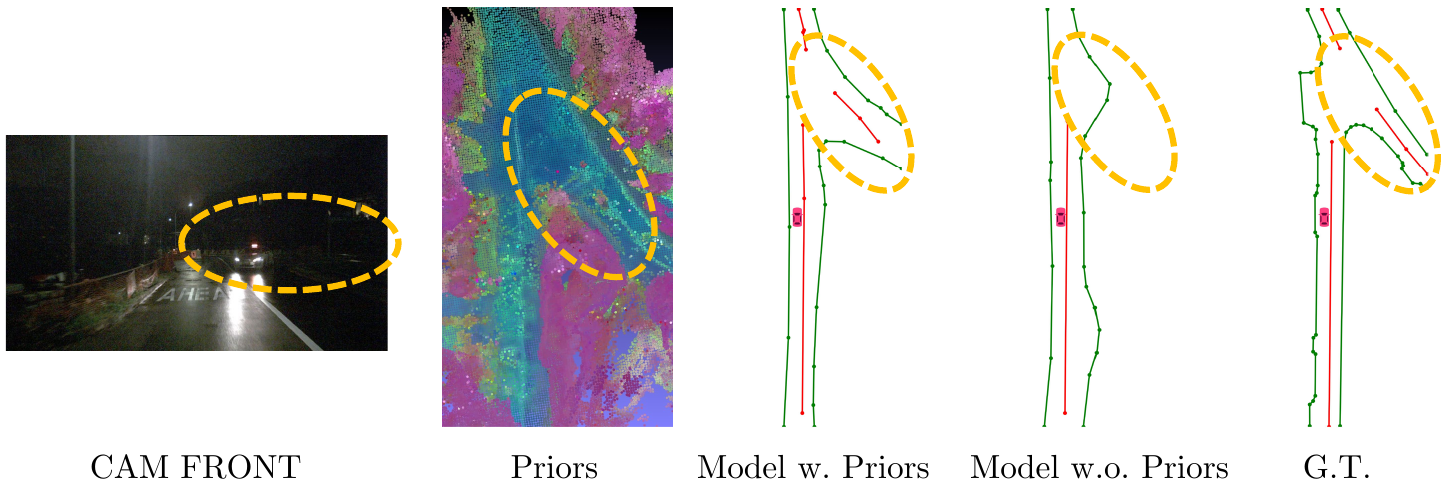}
  \caption{Visualization of StreamMapNet prediction with and without priors. The model with priors accurately predicts road structure, while baseline model without priors fails.
  }
  \label{fig:qualitative_map}
\end{figure}
\subsection{Qualitative Analysis}
In this section, we present qualitative results from our framework, underscoring the value of incorporating priors. Figure \ref{fig:qualitative_map} depicts a low-light scenario at an intersection that is hardly visible in the front-view image. Here, the baseline model without priors misinterprets the road layout, whereas the model equipped with priors accurately reconstructs it. The visualization of prior features reveals the detailed road structure, underscoring how priors provide crucial information for generating consistent and reliable HD maps critical for autonomous vehicle safety.

\begin{figure}[bh]
  \centering
  \includegraphics[width=1.0\linewidth]{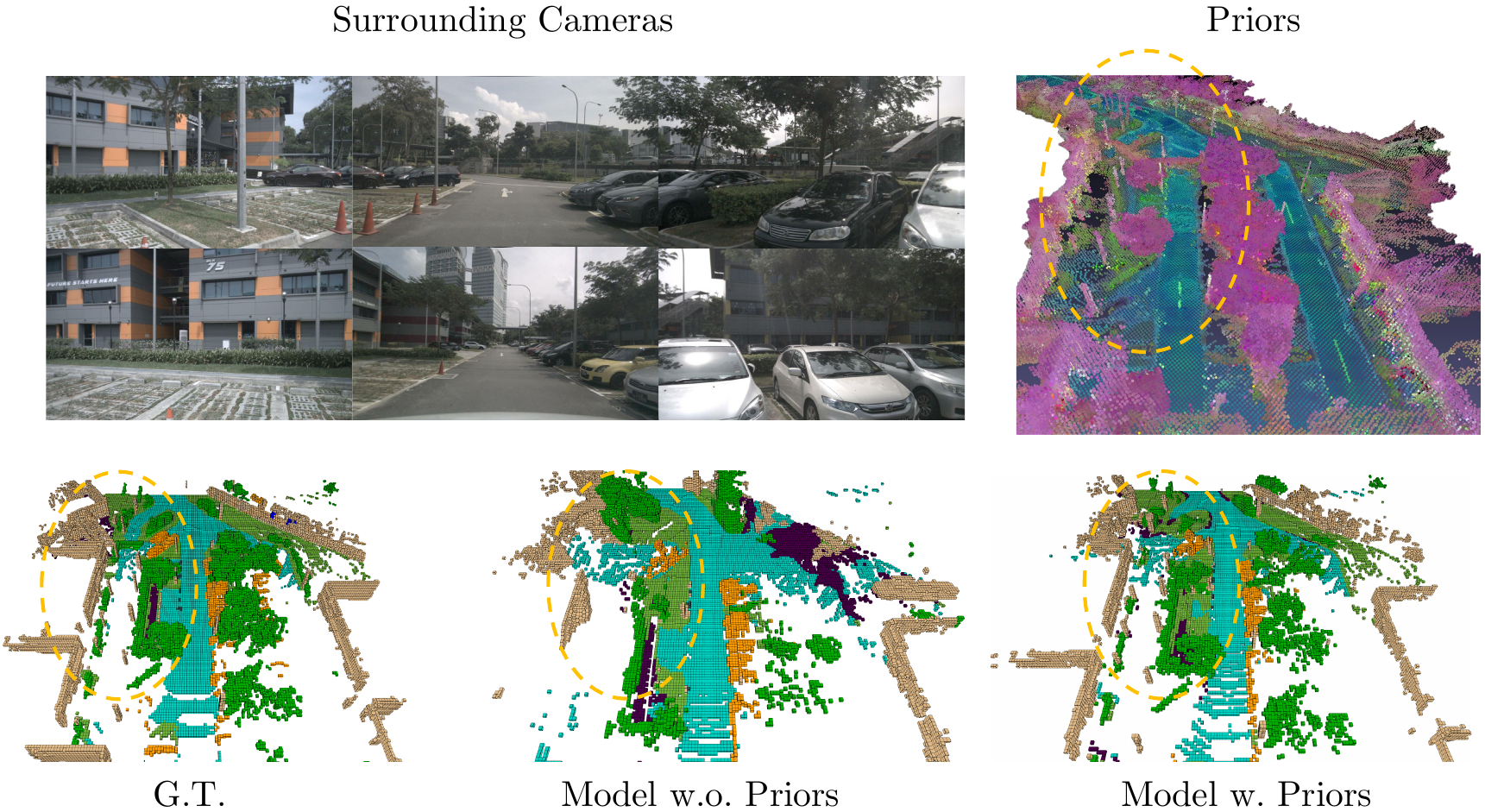}
  \caption{Visualization of BEVDet predictions and the pre-built priors. Self-supervised priors reconstruct fine-grained geometries even more denser than groundtruth, enabling the enhanced BEVDet model to deliver more precise occupancy predictions than baseline models lacking priors.
  }
  \label{fig:qualitative_occ}
\end{figure}
Figure \ref{fig:qualitative_occ} highlights the enhancement in occupancy prediction provided by our priors. Notably, within the highlighted region, our priors accurately capture the detailed structure of a light pole and the ground surface—details even absent in the ground truth. Consequently, the model with priors precisely predicts occupancy with fine details, in contrast to the baseline model's failure.

\section{Dicussion and Conclusion}
\noindent\textbf{Limitations.} Our approach relies on accurate ego-vehicle poses and camera sensors, usually provided by GPS and IMU systems, which might not be readily available in crowdsourced data. Future efforts could aim to reduce the prior memory size or exploring alternative self-supervised reconstruction methods like 3D Gaussian Splatting\cite{kerbl3Dgaussians}, offering promising directions for further research.

\noindent\textbf{Conclusion.} In this paper, we introduced PreSight, a novel framework that autonomously constructs static, city-scale priors for enhancing online perception in subsequent traversals. Our experiments on the nuScenes dataset demonstrate the framework's effectiveness and its wide applicability across various perception models. 


%
%

\input{main.bbl}
\newpage
\appendix

\section{Dataset}
As detailed in Section \ref{sec:setup}, we divide the nuScenes dataset into training, prior, and testing segments. Scenes from \textit{Boston Seaport} and \textit{Singapore Queenstown} comprise the training set, while \textit{Singapore Onenorth} and \textit{Singapore Holland Village} contribute to the prior and testing sets. To maintain an balanced distribution of day and night scenes, a subset of \textit{Singapore Holland Village}, including 32 night scenes, is allocated to the training set. This results in a division of 65 night scenes for training and 34 for testing, with night scenes excluded from prior construction. The distribution between prior and testing sets aims to balance overlap and scene diversity, yielding 19/34 scenes for prior/testing in \textit{Singapore Holland Village} and a 63/120 split in \textit{Singapore Onenorth}.
Here we list scene names of prior and testing set for reproduction.

\vspace{1ex}
\noindent\texttt{testing scenes: 0001, 0003, 0005, 0006, 0007, 0008, 0011, 0012, \\
0014, 0015, 0016, 0018, 0021, 0022, 0023, 0025, 0028, 0029, 0034, \\
0043, 0045, 0046, 0051, 0054, 0059, 0060, 0061, 0120, 0124, 0126, \\
0127, 0130, 0131, 0132, 0134, 0135, 0138, 0150, 0151, 0152, 0154, \\
0155, 0157, 0158, 0159, 0160, 0190, 0191, 0192, 0193, 0194, 0221, \\
0270, 0271, 0272, 0273, 0274, 0275, 0276, 0277, 0278, 0344, 0345, \\
0346, 0347, 0355, 0356, 0357, 0358, 0359, 0361, 0362, 0363, 0364, \\
0365, 0366, 0367, 0368, 0371, 0372, 0373, 0374, 0375, 0376, 0377, \\
0379, 0380, 0381, 0382, 0383, 0385, 0945, 0949, 0952, 0953, 0956, \\
0957, 0958, 0959, 0960, 0961, 0963, 0966, 0967, 0968, 0969, 0971, \\
0975, 0976, 0977, 0978, 0979, 0980, 0981, 0982, 0983, 0984, 0989, \\
0990, 0991, 1045, 1046, 1047, 1048, 1049, 1050, 1051, 1052, 1053, \\
1054, 1055, 1056, 1057, 1058, 1059, 1060, 1061, 1062, 1063, 1064, \\
1065, 1066, 1067, 1068, 1069, 1070, 1071, 1072, 1073, 1074, 1075, \\
1076, 1077, 1104.}\\
\noindent\texttt{prior scenes: 0002, 0004, 0009, 0010, 0013, 0017, 0019, 0020, \\
0024, 0026, 0027, 0030, 0031, 0032, 0033, 0035, 0036, 0038, 0039, \\
0041, 0042, 0044, 0047, 0048, 0049, 0050, 0052, 0053, 0055, 0056, \\
0057, 0058, 0121, 0122, 0123, 0125, 0128, 0129, 0133, 0139, 0149, \\
0195, 0196, 0268, 0269, 0348, 0349, 0350, 0351, 0352, 0353, 0354, \\
0360, 0369, 0370, 0378, 0384, 0386, 0399, 0400, 0401, 0402, 0403, \\
0405, 0406, 0407, 0408, 0410, 0411, 0412, 0413, 0414, 0415, 0416, \\
0417, 0418, 0419, 0947, 0955, 0962, 0972, 0988.}

\section{NeRF Implementation}
\noindent\textbf{Field Details.} Following the methodology of Mip-NeRF 360~\cite{barron2022mipnerf360}, our implementation adopts a two-stage proposal network for importance sampling, utilizing 128 and 64 sample points in the first and second stages, respectively. The proposal networks operate across resolutions from $2^4$ to $2^{10}$, featuring a maximum hash map capacity of $2^{20}$. These networks are structured into 8 levels, with each level encoding a 1-dimensional feature. For the main field, we define resolutions ranging from $2^4$ to $2^{14}$, maintaining the same maximum hash map capacity of $2^{20}$ but expanding to 10 levels, each with a 4-dimensional feature. All Multi-Layer Perceptrons (MLPs) in our framework consist of two hidden layers and are configured with a hidden dimension of 64.

\noindent\textbf{Optimization.} The comprehensive loss function is represented as:
\begin{equation}
    \mathcal{L} = \mathcal{L}_\mathrm{rgb} + \lambda_\mathrm{feat}\mathcal{L}_\mathrm{feat} + \lambda_\mathrm{sky}\mathcal{L}_\mathrm{sky} + \lambda_\mathrm{inter}\mathcal{L}_\mathrm{inter} + \lambda_\mathrm{dist}\mathcal{L}_\mathrm{dist}
\end{equation}
We set the weights for $\lambda_\mathrm{feat}$, $\lambda_\mathrm{sky}$, $\lambda_\mathrm{inter}$, and $\lambda_\mathrm{dist}$ at 0.5, 0.001, 1.0, and 0.002, respectively. For optimization, we employ the AdamW optimizer~\cite{loshchilov2018fixing} with an initial learning rate of 0.01, applying a decay factor of 0.33 at the $25\%$, $50\%$, and $75\%$ marks of the total iterations.

\section{Baseline Implementation}
This section outlines the implementation details for the baseline perception models, closely following their original codebases.

\noindent\textbf{HD Map Construction.} 
Following MapTR~\cite{MapTR} and StreamMapNet~\cite{Yuan_2024_streammapnet}, we adjust image resolution to $450\times800$ and use ResNet50~\cite{he2016deep} as the backbone. We use 24 training epochs, with learning rates set at $5\times 10^{-4}$ for MapTR and $6\times 10^{-4}$ for StreamMapNet, both with a batch size of 32, following their standard implementation. For the BEVFormer-based rasterized model, we use a batch size of 8 and a learning rate of $2\times 10^{-4}$, consistent with Neural Map Prior~\cite{xiong2023neuralmapprior}. Historical frames are not used for simplicity.

\noindent\textbf{Occupancy Prediction.} 
Following FB-OCC~\cite{li2023fbocc}, we resize images to $256\times 704$ and use ResNet50 as the image backbone. The models are trained for 24 epochs with BEV data augmentation. The learning rates are $2\times 10^{-4}$ for a batch size of 64 for FB-OCC and $1\times 10^{-4}$ for a batch size of 32 for BEVDet~\cite{huang2022bevdet4d}, in line with their official setups. Historical frames are incorporated to align with the original implementations.

\end{document}

%% file: tables/vector_map_1_runtime.tex
\begin{table}[tb]
  \caption{Our framework's enhancements across different online HD map construction models: StreamMapNet and MapTR use vectorized predictions, evaluated by Average Precision (AP), while BEVFormer's rasterized predictions are assessed using Intersection over Union (IoU).
  }
  \label{tab:vector_map_1}
  \centering
  \resizebox{0.9\textwidth}{!}{
      \setlength{\tabcolsep}{3pt}
      \begin{tabular}{cc|c|cccc|c}
    		\toprule
    		Model & Metric & w. Prior  & Ped Crossing & Divider & Boundary & All & Runtime (FPS)\\
    		\midrule
    		\multirow{2}{*}{StreamMapNet} & \multirow{2}{*}{AP} & $\times$ & 10.19 & 11.26 & 11.87 & 11.10 & \textbf{22.4}\\
             & & \checkmark & \textbf{21.11} & \textbf{23.73} &  \textbf{32.31}& \textbf{25.72 (+14.62)} & 21.9\\ 
             \midrule
            \multirow{2}{*}{MapTR} & \multirow{2}{*}{AP}  & $\times$ & 4.97 & 8.20 & 9.83 & 7.67 & \textbf{25.2} \\ 
            & & \checkmark & \textbf{16.18} & \textbf{19.04} & \textbf{34.14} & \textbf{23.12 (+15.45)} & 23.2\\ 
    		\midrule
            \multirow{2}{*}{BEVFormer} & \multirow{2}{*}{IoU}  & $\times$ & 14.90 & 29.88 & 32.74 & 25.84 & \textbf{15.5} \\ 
            & & \checkmark & \textbf{16.37} & \textbf{34.82} & \textbf{51.66} & \textbf{34.28 (+8.44)} & 14.3\\ 
            \bottomrule
      \end{tabular}
}
\end{table}

%% file: tables/occ_1_runtime.tex
\begin{table}[tb]
  \caption{Our framework boosts state-of-the-art occupancy prediction models, distinguishing between dynamic and static targets for clarity. The \textit{trailer} class is excluded due to its absence in our proposed testing set.
  }
  \label{tab:occ_1}
  \centering
  \resizebox{1.0\textwidth}{!}{
      \setlength{\tabcolsep}{2pt}
      \begin{tabular}{c|c|ccc|ccccc ccccc|cccc cc|c}
        \toprule
        &&&&&\multicolumn{10}{c|}{Dynamic} & \multicolumn{6}{c|}{\textbf{Static}}& \\
        \cline{6-21}
        Method & \rotatebox[origin=lb]{90}{w. Priors} 
        & \rotatebox[origin=lb]{90}{mIoU}
        & \rotatebox[origin=lb]{90}{Dynamic}
        & \rotatebox[origin=lb]{90}{\textbf{Static}}
        & \rotatebox[origin=lb]{90}{others}
        & \rotatebox[origin=lb]{90}{barrier}
        & \rotatebox[origin=lb]{90}{bicycle}
        & \rotatebox[origin=lb]{90}{bus}
        & \rotatebox[origin=lb]{90}{car}
        & \rotatebox[origin=lb]{90}{constr. vehicle\ }
        & \rotatebox[origin=lb]{90}{motorcycle}
        & \rotatebox[origin=lb]{90}{pedestrian}
        & \rotatebox[origin=lb]{90}{traffic cone}
        & \rotatebox[origin=lb]{90}{truck}
        & \rotatebox[origin=lb]{90}{drive surface}
        & \rotatebox[origin=lb]{90}{other flat}
        & \rotatebox[origin=lb]{90}{sidewalk}
        & \rotatebox[origin=lb]{90}{terrain}
        & \rotatebox[origin=lb]{90}{manmade}
        & \rotatebox[origin=lb]{90}{vegetation}
        & \rotatebox[origin=lb]{90}{Runtime (FPS)} 
        \\
        \midrule
        \multirow{2}{*}{BEVDet} & $\times$ & 29.3 & 24.4 & 38.2 & \textbf{1.5} & \textbf{42.4} & 11.0 & \textbf{43.0} & 47.1 & \textbf{19.1} & 23.3 & 23.4 & \textbf{19.5} & \textbf{37.8} & 72.9 & 11.6 & 30.9 & 48.6 & 32.7 & 32.5 & \textbf{5.1}\\
        & \checkmark & \textbf{33.7} & \textbf{24.4} & \textbf{50.5} & 1.2 & 40.1 & \textbf{14.8} & 42.1 & \textbf{48.3} & 15.7 & \textbf{26.4} & \textbf{24.4} & 18.7 & 37.2 & \textbf{81.8} & \textbf{15.2} & \textbf{40.3} & \textbf{60.5} & \textbf{50.4} & \textbf{54.9} & 4.9\\
        \midrule
        \multirow{2}{*}{FB-Occ} & $\times$ & 30.0 & 25.1 & 39.2 & 9.2 & 37.2 & \textbf{21.8} & \textbf{41.6} & 43.4 & 15.8 & 27.3 & 25.4 & \textbf{23.8} & \textbf{30.3} & 74.7 & 17.3 & 33.0 & 50.6 & 28.2 & 31.1 & \textbf{9.1} \\
        & \checkmark & \textbf{34.3} & \textbf{25.4} & \textbf{50.7} & \textbf{9.3} & \textbf{38.3} & 21.0 & 40.3 & \textbf{45.0} & \textbf{15.9}& \textbf{29.9} & \textbf{26.0} & 23.8 & 30.2 & \textbf{82.3} & \textbf{18.5} & \textbf{39.1} & \textbf{61.2} & \textbf{48.0} & \textbf{54.7} & 8.6\\
        \bottomrule
      \end{tabular}
  }
\end{table}

%% file: tables/compare.tex
\begin{table}[tb]
  \caption{Our method outperforms other priors in HD map construction and achieves similar results to LiDAR priors in occupancy prediction, using only camera inputs.
  }
  \label{tab:compare}
  \vspace{-1ex}
  \centering
  \begin{subtable}{.5\linewidth}
      \centering
        \caption{HD Map Construction}
        \setlength{\tabcolsep}{5pt}
        \resizebox{.95\textwidth}{!} {
            \begin{tabular}{cc|c|c}
                \toprule
                Model & Metric & Prior Type & mAP/mIoU\\
                \midrule
                \multirow{3}{*}{StreamMapNet} & \multirow{3}{*}{AP} & 
                $\times$ & 11.10 \\
                & & LiDAR & 16.87 \\
                & & \textbf{ours} & \textbf{25.72}\\
                \midrule
                \multirow{3}{*}{MapTR} & \multirow{3}{*}{AP} & 
                $\times$ & 7.67 \\
                & & LiDAR & 15.93 \\
                & & \textbf{ours} & \textbf{23.12}\\
                \midrule
                \multirow{4}{*}{BEVFormer} & \multirow{4}{*}{IoU} & 
                $\times$ & 25.84 \\
                & & LiDAR & 33.01 \\
                & & NMP & 26.50 \\
                & & \textbf{ours} & \textbf{34.42}\\
                \bottomrule
            \end{tabular}
        }
  \end{subtable}%
  \begin{subtable}{.5\linewidth}
      \centering
        \caption{Occupancy Prediction}
        \setlength{\tabcolsep}{5pt}
        \resizebox{.95\textwidth}{!} {
            \begin{tabular}{c|c|ccc}
                \toprule
                Model & Prior Type & mIoU & Dynamic & Static\\
                \midrule
                \multirow{3}{*}{BEVDet} & 
                $\times$ & 29.3 & 24.4 & 38.2 \\
                & LiDAR & 33.1 & 21.7 & \textbf{53.9} \\
                & \textbf{ours} & \textbf{33.7} & \textbf{24.4} & 50.5\\
                \midrule
                \multirow{3}{*}{FB-OCC} & 
                $\times$ & 30.0 & 25.1 & 39.2 \\
                & LiDAR & 34.3 & 23.4 & \textbf{54.4} \\
                & \textbf{ours} & \textbf{34.3} & \textbf{25.4} & 50.7\\
                \bottomrule
            \end{tabular}
        }
  \end{subtable} 
\end{table}

%% file: tables/nofeats.tex
\begin{table}[tb]
  \caption{Ablation study reveals the impact of distilled semantic features: Priors enhance online perception on their own, but performance significantly improves with the addition of distilled features. 
  }
  \label{tab:no_feats}
  \vspace{-1ex}
  \centering
  \begin{subtable}{.5\linewidth}
      \centering
        \caption{HD Map Construction}
        \setlength{\tabcolsep}{5pt}
        \resizebox{.8\textwidth}{!} {
            \begin{tabular}{c|c|c}
                \toprule
                Model & Prior Type & mAP\\
                \midrule
                \multirow{3}{*}{StreamMapNet} & 
                $\times$ & 11.10 \\
                & no feats. & 11.45 \\
                & \textbf{full} & \textbf{25.72}\\
                \bottomrule
            \end{tabular}
        }
  \end{subtable}%
  \begin{subtable}{.5\linewidth}
      \centering
        \caption{Occupancy Prediction}
        \setlength{\tabcolsep}{5pt}
        \resizebox{1.0\textwidth}{!} {
            \begin{tabular}{c|c|ccc}
                \toprule
                Model & Prior Type & mIoU & Dynamic & Static\\
                \midrule
                \multirow{3}{*}{BEVDet} & 
                $\times$ & 29.3 & 24.4 & 38.2 \\
                & no feats. & 32.0 & 24.1 & 46.4 \\
                & \textbf{full} & \textbf{33.7} & \textbf{24.4} & \textbf{50.5}\\
                \bottomrule
            \end{tabular}
        }
  \end{subtable} 
\end{table}

%% file: tables/prior_ratio.tex
\begin{table}[tb]
  \caption{Performance improves as as we scales up priors scenes. $\times$ represents baseline models. $0\%$ denotes that no priors are used during testing, whereas $50\%$ indicates that only half of the scenes in $X_{prior}$ are used for testing. A $100\%$ ratio means all scenes in $X_{prior}$ are used.
  }
  \label{tab:prior_ratio}
  \vspace{-1ex}
  \centering
  \begin{subtable}{.5\linewidth}
      \centering
        \caption{HD Map Construction}
        \setlength{\tabcolsep}{5pt}
        \resizebox{.8\textwidth}{!} {
            \begin{tabular}{c|c|c}
                \toprule
                Model & Prior Type & mAP\\
                \midrule
                \multirow{3}{*}{StreamMapNet} & 
                $\times$ & 11.10 \\
                & $0\%$ & 12.03 \\
                & $50\%$ & 21.62 \\
                & $\mathbf{100\%}$ & \textbf{25.72}\\
                \bottomrule
            \end{tabular}
        }
  \end{subtable}%
  \begin{subtable}{.5\linewidth}
      \centering
        \caption{Occupancy Prediction}
        \setlength{\tabcolsep}{5pt}
        \resizebox{1.0\textwidth}{!} {
            \begin{tabular}{c|c|ccc}
                \toprule
                Model & Prior Type & mIoU & Dynamic & Static\\
                \midrule
                \multirow{3}{*}{BEVDet} & 
                $\times$ & 29.3 & 24.4 & 38.2 \\
                & $0\%$ & 28.7 & 24.6 & 36.2 \\
                & $50\%$ & 32.0 & 24.1 & 46.4 \\
                & $\mathbf{100\%}$ & \textbf{33.7} & \textbf{24.4} & \textbf{50.5}\\
                \bottomrule
            \end{tabular}
        }
  \end{subtable} 
\end{table}

%% file: tables/localization.tex
\begin{table}[tb]
  \caption{Performance for online models with noisy localization.
  }
  \label{tab:localization}
  \centering
  \resizebox{0.6\textwidth}{!}{
      \setlength{\tabcolsep}{3pt}
      \begin{tabular}{c|ccc}
    		\toprule
            &\multicolumn{3}{|c}{mAP/mIoU}\\
    		Model & w.o. Noise & w. Noise & w.o. Priors \\
            \midrule
            StreamMapNet & 25.7 & 24.5 & 11.1 \\
            BEVDet & 33.7 & 32.5 & 29.3\\
            \bottomrule
      \end{tabular}
    }
\end{table}

%% file: tables/prior_only.tex
\begin{table}[tb]
  \caption{Experiments on priors-only models. "Setting a" evaluates models trained with both online observations and priors, replacing the online BEV features with zeros. "Setting b" involves training new models using only priors during both training and testing. "Full" represents our best models. "Baseline" denotes models without any priors.}
  \label{tab:prior_only}
  \centering
  \resizebox{0.6\textwidth}{!}{
      \setlength{\tabcolsep}{3pt}
      \begin{tabular}{c|cccc}
    		\toprule
            &\multicolumn{4}{|c}{mAP}\\
    		Model & Setting a & Setting b & Full & Baseline \\
            \midrule
            StreamMapNet & 1.9 & 19.8 & 25.7 & 11.1 \\
            MapTR & 6.0 & 18.8 & 23.1 & 7.7 \\
            \bottomrule
      \end{tabular}
    }
\end{table}

%% file: main.bbl
\begin{thebibliography}{10}
\providecommand{\url}[1]{\texttt{#1}}
\providecommand{\urlprefix}{URL }
\providecommand{\doi}[1]{https://doi.org/#1}

\bibitem{2012arie_global}
Arie-Nachimson, M., Kovalsky, S.Z., Kemelmacher-Shlizerman, I., Singer, A., Basri, R.: Global motion estimation from point matches. In: 2012 Second International Conference on 3D Imaging, Modeling, Processing, Visualization \& Transmission. pp. 81--88 (2012). \doi{10.1109/3DIMPVT.2012.46}

\bibitem{Aulinas2008slam}
Aulinas, J., Petillot, Y., Salvi, J., Llado, X.: The slam problem: a survey. vol.~184, pp. 363--371 (01 2008). \doi{10.3233/978-1-58603-925-7-363}

\bibitem{barron2022mipnerf360}
Barron, J.T., Mildenhall, B., Verbin, D., Srinivasan, P.P., Hedman, P.: Mip-nerf 360: Unbounded anti-aliased neural radiance fields. CVPR  (2022)

\bibitem{caesar2020nuscenes}
Caesar, H., Bankiti, V., Lang, A.H., Vora, S., Liong, V.E., Xu, Q., Krishnan, A., Pan, Y., Baldan, G., Beijbom, O.: nuscenes: A multimodal dataset for autonomous driving. In: Proceedings of the IEEE/CVF conference on computer vision and pattern recognition. pp. 11621--11631 (2020)

\bibitem{caron2021emerging}
Caron, M., Touvron, H., Misra, I., J\'egou, H., Mairal, J., Bojanowski, P., Joulin, A.: Emerging properties in self-supervised vision transformers. In: Proceedings of the International Conference on Computer Vision (ICCV) (2021)

\bibitem{Crandall2011Discrete}
Crandall, D., Owens, A., Snavely, N., Huttenlocher, D.: Discrete-continuous optimization for large-scale structure from motion. In: CVPR 2011. pp. 3001--3008 (2011). \doi{10.1109/CVPR.2011.5995626}

\bibitem{dellaert2012factor}
Dellaert, F.: Factor graphs and gtsam: A hands-on introduction. Tech. rep., Georgia Institute of Technology (2012)

\bibitem{ding2023pivotnet}
Ding, W., Qiao, L., Qiu, X., Zhang, C.: Pivotnet: Vectorized pivot learning for end-to-end hd map construction. In: Proceedings of the IEEE/CVF International Conference on Computer Vision. pp. 3672--3682 (2023)

\bibitem{ransac}
Fischler, M.A., Bolles, R.C.: Random sample consensus: a paradigm for model fitting with applications to image analysis and automated cartography. Commun. ACM  \textbf{24}(6),  381–395 (jun 1981). \doi{10.1145/358669.358692}, \url{https://doi.org/10.1145/358669.358692}

\bibitem{Fuentes2015vslam}
Fuentes-Pacheco, J., Ascencio, J., Rendon-Mancha, J.: Visual simultaneous localization and mapping: A survey. Artificial Intelligence Review  \textbf{43} (11 2015). \doi{10.1007/s10462-012-9365-8}

\bibitem{Goldstein2016ShapeFit}
Goldstein, T., Hand, P., Lee, C., Voroninski, V., Soatto, S.: Shapefit and shapekick for robust, scalable structure from motion  (08 2016)

\bibitem{guo2023streetsurf}
Guo, J., Deng, N., Li, X., Bai, Y., Shi, B., Wang, C., Ding, C., Wang, D., Li, Y.: Streetsurf: Extending multi-view implicit surface reconstruction to street views. arXiv preprint arXiv:2306.04988  (2023)

\bibitem{he2016deep}
He, K., Zhang, X., Ren, S., Sun, J.: Deep residual learning for image recognition. In: Proceedings of the IEEE conference on computer vision and pattern recognition. pp. 770--778 (2016)

\bibitem{hong2024univision}
Hong, Y., Liu, Q., Cheng, H., Ma, D., Dai, H., Wang, Y., Cao, G., Ding, Y.: Univision: A unified framework for vision-centric 3d perception (2024)

\bibitem{huang2022bevdet4d}
Huang, J., Huang, G.: Bevdet4d: Exploit temporal cues in multi-camera 3d object detection. arXiv preprint arXiv:2203.17054  (2022)

\bibitem{huang2021bevdet}
Huang, J., Huang, G., Zhu, Z., Yun, Y., Du, D.: Bevdet: High-performance multi-camera 3d object detection in bird-eye-view. arXiv preprint arXiv:2112.11790  (2021)

\bibitem{huang2023self}
Huang, Y., Zheng, W., Zhang, B., Zhou, J., Lu, J.: Selfocc: Self-supervised vision-based 3d occupancy prediction. arXiv preprint arXiv:2311.12754  (2023)

\bibitem{huang2023tri}
Huang, Y., Zheng, W., Zhang, Y., Zhou, J., Lu, J.: Tri-perspective view for vision-based 3d semantic occupancy prediction. arXiv preprint arXiv:2302.07817  (2023)

\bibitem{kerbl3Dgaussians}
Kerbl, B., Kopanas, G., Leimk{\"u}hler, T., Drettakis, G.: 3d gaussian splatting for real-time radiance field rendering. ACM Transactions on Graphics  \textbf{42}(4) (July 2023), \url{https://repo-sam.inria.fr/fungraph/3d-gaussian-splatting/}

\bibitem{li2021hdmapnet}
Li, Q., Wang, Y., Wang, Y., Zhao, H.: Hdmapnet: A local semantic map learning and evaluation framework. arXiv preprint arXiv:2107.06307  (2021)

\bibitem{li2023voxformer}
Li, Y., Yu, Z., Choy, C., Xiao, C., Alvarez, J.M., Fidler, S., Feng, C., Anandkumar, A.: Voxformer: Sparse voxel transformer for camera-based 3d semantic scene completion. In: Proceedings of the IEEE/CVF Conference on Computer Vision and Pattern Recognition (CVPR) (2023)

\bibitem{li2022bevformer}
Li, Z., Wang, W., Li, H., Xie, E., Sima, C., Lu, T., Qiao, Y., Dai, J.: Bevformer: Learning bird’s-eye-view representation from multi-camera images via spatiotemporal transformers. arXiv preprint arXiv:2203.17270  (2022)

\bibitem{li2023fbocc}
Li, Z., Yu, Z., Austin, D., Fang, M., Lan, S., Kautz, J., Alvarez, J.M.: {FB-OCC}: {3D} occupancy prediction based on forward-backward view transformation. arXiv:2307.01492  (2023)

\bibitem{li2023fbbev}
Li, Z., Yu, Z., Wang, W., Anandkumar, A., Lu, T., Alvarez, J.M.: {FB-BEV}: {BEV} representation from forward-backward view transformations. In: IEEE/CVF International Conference on Computer Vision (ICCV) (2023)

\bibitem{li2023voxelformer}
Li, Z., Zhang, C., Ma, W.C., Zhou, Y., Huang, L., Wang, H., Lim, S., Zhao, H.: Voxelformer: Bird's-eye-view feature generation based on dual-view attention for multi-view 3d object detection (2023)

\bibitem{MapTR}
Liao, B., Chen, S., Wang, X., Cheng, T., Zhang, Q., Liu, W., Huang, C.: Maptr: Structured modeling and learning for online vectorized hd map construction. In: International Conference on Learning Representations (2023)

\bibitem{lilja2023localization}
Lilja, A., Fu, J., Stenborg, E., Hammarstrand, L.: Localization is all you evaluate: Data leakage in online mapping datasets and how to fix it (2023)

\bibitem{lin2023sparse4d}
Lin, X., Lin, T., Pei, Z., Huang, L., Su, Z.: Sparse4d v2: Recurrent temporal fusion with sparse model (2023)

\bibitem{liu2023vectormapnet}
Liu, Y., Yuan, T., Wang, Y., Wang, Y., Zhao, H.: Vectormapnet: End-to-end vectorized hd map learning (2023)

\bibitem{liu2022bevfusion}
Liu, Z., Tang, H., Amini, A., Yang, X., Mao, H., Rus, D., Han, S.: Bevfusion: Multi-task multi-sensor fusion with unified bird's-eye view representation. In: IEEE International Conference on Robotics and Automation (ICRA) (2023)

\bibitem{loshchilov2018fixing}
Loshchilov, I., Hutter, F.: Fixing weight decay regularization in adam  (2018)

\bibitem{lowe2004distinctive}
Lowe, D.G.: Distinctive image features from scale-invariant keypoints. International journal of computer vision  \textbf{60},  91--110 (2004)

\bibitem{mildenhall2020nerf}
Mildenhall, B., Srinivasan, P.P., Tancik, M., Barron, J.T., Ramamoorthi, R., Ng, R.: Nerf: Representing scenes as neural radiance fields for view synthesis. In: ECCV (2020)

\bibitem{mueller2022instant}
M\"uller, T., Evans, A., Schied, C., Keller, A.: Instant neural graphics primitives with a multiresolution hash encoding. ACM Trans. Graph.  \textbf{41}(4),  102:1--102:15 (Jul 2022). \doi{10.1145/3528223.3530127}, \url{https://doi.org/10.1145/3528223.3530127}

\bibitem{roddick2020predicting}
Roddick, T., Cipolla, R.: Predicting semantic map representations from images using pyramid occupancy networks. In: Proceedings of the IEEE/CVF Conference on Computer Vision and Pattern Recognition. pp. 11138--11147 (2020)

\bibitem{segal2009generalized}
Segal, A., Hähnel, D., Thrun, S.: Generalized-icp. In: Robotics: science and systems (06 2009). \doi{10.15607/RSS.2009.V.021}

\bibitem{legoloam2018}
Shan, T., Englot, B.: Lego-loam: Lightweight and ground-optimized lidar odometry and mapping on variable terrain. In: IEEE/RSJ International Conference on Intelligent Robots and Systems (IROS). pp. 4758--4765. IEEE (2018)

\bibitem{phototourism}
Snavely, N., Seitz, S.M., Szeliski, R.: Photo tourism: exploring photo collections in 3D. Association for Computing Machinery, New York, NY, USA, 1 edn. (2023), \url{https://doi.org/10.1145/3596711.3596766}

\bibitem{tancik2022blocknerf}
Tancik, M., Casser, V., Yan, X., Pradhan, S., Mildenhall, B., Srinivasan, P.P., Barron, J.T., Kretzschmar, H.: Block-nerf: Scalable large scene neural view synthesis  (Feb 2022), \url{http://arxiv.org/abs/2202.05263v1}

\bibitem{tian2023occ3d}
Tian, X., Jiang, T., Yun, L., Wang, Y., Wang, Y., Zhao, H.: Occ3d: A large-scale 3d occupancy prediction benchmark for autonomous driving. arXiv preprint arXiv:2304.14365  (2023)

\bibitem{neurad}
Tonderski, A., Lindstr{\"o}m, C., Hess, G., Ljungbergh, W., Svensson, L., Petersson, C.: Neurad: Neural rendering for autonomous driving. arXiv preprint arXiv:2311.15260  (2023)

\bibitem{turki2023suds}
Turki, H., Zhang, J.Y., Ferroni, F., Ramanan, D.: Suds: Scalable urban dynamic scenes. In: Computer Vision and Pattern Recognition (CVPR) (2023)

\bibitem{wang2021fcos3d}
Wang, T., Zhu, X., Pang, J., Lin, D.: {FCOS3D: Fully} convolutional one-stage monocular 3d object detection. In: Proceedings of the IEEE/CVF International Conference on Computer Vision (ICCV) Workshops (2021)

\bibitem{wang2022detr3d}
Wang, Y., Guizilini, V.C., Zhang, T., Wang, Y., Zhao, H., Solomon, J.: Detr3d: 3d object detection from multi-view images via 3d-to-2d queries. In: Conference on Robot Learning. pp. 180--191. PMLR (2022)

\bibitem{xie2021segformer}
Xie, E., Wang, W., Yu, Z., Anandkumar, A., Alvarez, J.M., Luo, P.: Segformer: Simple and efficient design for semantic segmentation with transformers. In: Neural Information Processing Systems (NeurIPS) (2021)

\bibitem{xie2023mvmap}
Xie, Z., Pang, Z., Wang, Y.X.: Mv-map: Offboard hd-map generation with multi-view consistency. arXiv  (2023)

\bibitem{xiong2023neuralmapprior}
Xiong, X., Liu, Y., Yuan, T., Wang, Y., Wang, Y., Hang, Z.: Neural map prior for autonomous driving. In: CVPR (2023)

\bibitem{Yang2022BEVFormerVA}
Yang, C., Chen, Y., Tian, H., Tao, C., Zhu, X., Zhang, Z., Huang, G., Li, H., Qiao, Y., Lu, L., Zhou, J., Dai, J.: Bevformer v2: Adapting modern image backbones to bird's-eye-view recognition via perspective supervision. ArXiv  (2022)

\bibitem{yang2023emernerf}
Yang, J., Ivanovic, B., Litany, O., Weng, X., Kim, S.W., Li, B., Che, T., Xu, D., Fidler, S., Pavone, M., Wang, Y.: Emernerf: Emergent spatial-temporal scene decomposition via self-supervision. arXiv preprint arXiv:2311.02077  (2023)

\bibitem{yang2023unisim}
Yang, Z., Chen, Y., Wang, J., Manivasagam, S., Ma, W.C., Yang, A.J., Urtasun, R.: Unisim: A neural closed-loop sensor simulator. In: CVPR (2023)

\bibitem{you2022hindsight}
You, Y., Luo, K.Z., Chen, X., Chen, J., Chao, W.L., Sun, W., Hariharan, B., Campbell, M., Weinberger, K.Q.: Hindsight is 20/20: Leveraging past traversals to aid 3d perception. arXiv preprint arXiv:2203.11405  (2022)

\bibitem{Yuan_2024_streammapnet}
Yuan, T., Liu, Y., Wang, Y., Wang, Y., Zhao, H.: Streammapnet: Streaming mapping network for vectorized online hd map construction. In: Proceedings of the IEEE/CVF Winter Conference on Applications of Computer Vision (WACV). pp. 7356--7365 (January 2024)

\bibitem{chubin2023occnerf}
Zhang, C., Yan, J., Wei, Y., Li, J., Liu, L., Tang, Y., Duan, Y., Lu, J.: Occnerf: Self-supervised multi-camera occupancy prediction with neural radiance fields. arXiv preprint arXiv:2312.09243  (2023)

\bibitem{loamzhang2014}
Zhang, J., Singh, S.: Loam : Lidar odometry and mapping in real-time. Robotics: Science and Systems Conference (RSS) pp. 109--111 (01 2014)

\bibitem{Özyeşil_Voroninski_Basri_Singer_2017}
Özyeşil, O., Voroninski, V., Basri, R., Singer, A.: A survey of structure from motion. Acta Numerica  \textbf{26},  305–364 (2017). \doi{10.1017/S096249291700006X}

\end{thebibliography}
